%% file: arxiv_main.tex
\documentclass[10pt,twocolumn,letterpaper]{article}

\usepackage[pagenumbers]{wacv} % To force page numbers, e.g. for an arXiv version

\usepackage{graphicx}
\usepackage{amsmath}
\usepackage{amssymb}
\usepackage{booktabs}
\usepackage{placeins}

\usepackage[pagebackref,breaklinks,colorlinks]{hyperref}

\usepackage[capitalize]{cleveref}
\crefname{section}{Sec.}{Secs.}
\Crefname{section}{Section}{Sections}
\Crefname{table}{Table}{Tables}
\crefname{table}{Tab.}{Tabs.}

\def\wacvPaperID{285} % *** Enter the WACV Paper ID here
\def\confName{WACV}
\def\confYear{2025}

\newcommand{\vect}[1]{\boldsymbol{\mathbf{#1}}}

\begin{document}

\title{NPL-MVPS: Neural Point-Light Multi-View Photometric Stereo}

%\author{Fotios Logothetis\inst{1} \and
%Ignas Budvytis\inst{1} \and
%Roberto Cipolla \inst{1,2}
%\authorrunning{F.~Logothetis et al.}

%\institute{Toshiba Europe \\
%email{fotios.logothetis@toshiba.eu}\\
% \and
%University of Cambridge\\
%\email{rc10001@cam.ac.uk}}

\author{Fotios Logothetis\\
Toshiba Europe \\
Cambridge UK\\
{\tt\small fotios.logothetis@toshiba.eu}
\and
Ignas Budvytis\\
Independent researcher\\
Cambridge UK\\
{\tt\small ignas.budvytis@gmail.com}\\
~
\and
Roberto Cipolla \\
University of Cambridge\\
Cambridge UK\\
{\tt\small rc10001@cam.ac.uk}
}
\maketitle

%%%%%%%%% ABSTRACT
\begin{abstract}
In this work we present a novel multi-view photometric stereo (MVPS) method. Like many works in 3D reconstruction we are leveraging neural shape representations and learnt renderers. However, our work differs from the state-of-the-art multi-view PS methods such as PS-NeRF~\cite{psnerf} or Supernormal~\cite{cao2023supernormal} in that we explicitly leverage per-pixel intensity renderings rather than relying mainly on estimated normals.

We model point light attenuation and explicitly raytrace cast shadows in order to best approximate the incoming radiance for each point. The estimated incoming radiance is used as input to a fully neural material renderer that uses minimal prior assumptions and it is jointly optimised with the surface. Estimated normals and segmentation maps are also incorporated in order to maximise the surface accuracy.

Our method is among the first (along with Supernormal~\cite{cao2023supernormal}) to outperform the classical MVPS approach proposed by the DiLiGenT-MV benchmark and achieves average 0.2mm Chamfer distance for objects imaged at approx 1.5m  distance away with approximate $400\times400$ resolution. Moreover, our method shows high robustness to the sparse MVPS setup (6 views, 6 lights) greatly outperforming the SOTA competitor (0.38mm vs 0.61mm), illustrating the importance of neural rendering in multi-view photometric stereo.  %to poor normals in low light count scenario, achieving 0.27mm Chamfer distance when pixel rendering is used instead of estimated normals.

%\textbf{key point: more correct rendering but has limitations (i.e. no self reflection) so we need to also use normals}

\end{abstract}

\input{sections/introduction}    
\input{sections/relatedworks}    
\input{sections/method}
\input{sections/experimentsetup}  
\input{sections/experiments}
\input{sections/conclusions}

\FloatBarrier
\newpage
%%%%%%%%% REFERENCES
{\small
\bibliographystyle{ieee_fullname}
\bibliography{egbib}
}
\appendix

\section{Appendix}

 This appendix contains qualitative results on DiLiGenT-MV~\cite{LiZWSDT20} benchmark in Section~\ref{sec:sparse_results} and a brief discussion of DiLiGenT-MV data anomalies in Section~\ref{sec:dili_prob}.

\input{sections/appendix}

\end{document}

%% file: sections/introduction.tex
\section{Introduction}
\label{sec:intro}

Photometric Stereo (PS) is a long standing and important problem in the field of Computer Vision. While early PS works~\cite{Woodham89,logothetis2016near,ikehata2018cnn,Logothetis22,ikehata2023sdmunips} primarily tackled the estimation of normals from  single view images, the value of PS was unlocked by binocular~\cite{KongXT06,DuGS11,WangWMHCS13,Logothetis24WACV} and multi-view~\cite{EstebanVC08,zhou13mvps,park17mvps,Logothetis19,LiZWSDT20,psnerf} stereo setups as it allowed for accurate recovery of shape and not only normals. This, in turn, opened many applications such as general 3D reconstruction, novel-view rendering, relighting and material editing~\cite{psnerf}, as well as robot interaction, quality control in manufacturing and industrial conveyor belt scanning. 

\begin{figure*}[ht]
\includegraphics[width=\textwidth]{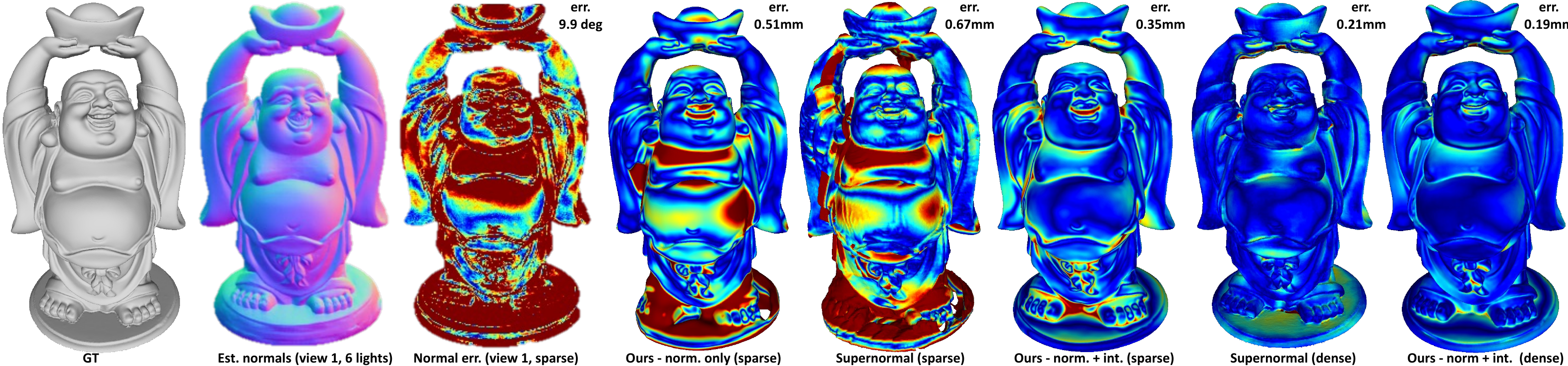}
\caption{In this figure we demonstrate the fragility of relying mainly on estimated normals (using~\cite{hardyunips}) for deep learning based \textit{sparse} multi-view photometric stereo when 6 views out of 20 and 6 light sources out of 96 available are used. The first column shows the ground truth of Buddha object from \textit{dense} MVPS DiLiGenT-MV benchmark. The following two columns show the estimated normals and corresponding error maps using only 6 out of 96 lights available (view 1 mean average normal error is $9.9^{\circ}$, saturated red color corresponds to $5^{\circ}$ error). Using such normals leads to a large reconstruction error using our method when pixel intensities are not leveraged (0.51mm) and previous SOTA \textit{dense} MVPS method Supernormal~\cite{cao2023supernormal} (0.67mm). If pixel intensities are used along with estimated normals (column 6) a significantly smaller error of 0.35mm is achieved. The final two columns show the error maps of estimated shapes when all available views and lights are used. In this setting Supernormal~\cite{cao2023supernormal} achieves a similar reconstruction error as our method (0.21mm vs 0.19mm). Similar dynamics apply to other DiLiGenT-MV objects as shown in Tables~\ref{tab:Tab_eval_diligent} and \ref{tab:Tab_eval_diligent_sparse}, strongly motivating for explicit pixel intensity modeling in MVPS methods. Note here the errors are computed  as Chamfer distance while the visualisation only shows errors from reconstruction to ground truth mesh for each reconstructed mesh surface point. Note dark red corresponds to $\ge 1mm$ error in the shape error illustrations (columns 3-8).}
\label{fig:intro}
\end{figure*}

Along with increasing the number of views Photometric Stereo undergone another important change by moving from classical non-linear optimisation enabled inverse graphics approaches (for single view~\cite{hui2017shape}, binocular~\cite{KongXT06}, multi-view~\cite{Logothetis19,LiZWSDT20}) to neural network (e.g.~\cite{ikehata2018cnn,logothetis2021pxnet}) and in particular neural shape representation enabled inverse graphics approaches (for single view~\cite{guo2022edgepreserving}, binocular~\cite{Logothetis24WACV} and multi-view~\cite{psnerf,cao2023supernormal}). However, despite the latter methods, especially~\cite{cao2023supernormal}, achieving impressive accuracy on DiLiGenT-MV~\cite{LiZWSDT20} benchmark their approach to MVPS is somewhat incomplete as they do not attempt to directly explain (and learn to match) observed pixel-wise intensities. In particular, ~\cite{cao2023supernormal} does not explicitly use image intensities to optimise for shape and is fully reliant on per-view  normal maps. Whereas, PS-NeRF~\cite{psnerf} only uses  average intensity during the surface optimisation stage and thus leaves most of the photometric information unused. It is important to note that Brahimi et. al. \cite{Brahimi_2024_CVPR} , attempts to re-render the images however does not model cast shadows and uses the simplified Dinsey BRDF \cite{burley2012physically} which may not model all materials accurately.

%\textcolor{red}{[EXPLAIN WHY BRAHIMI IS GOOD APPROACH BUT DOES NOT ACHIEVE GOOD RESULTS]}

%While it is widely believed that neural versions of the PS are superior to classical ones we discover that more than 10 years after the dawn of the latest era in Deep Learning~\cite{alexnet} multi-view photometric stereo (MVPS) current best solution still remains a classical one, proposed by the authors of the DiLiGenT-MV~\cite{LiZWSDT20} benchmark author as opposed to neural representation enabled PS-NeRF~\cite{psnerf} despite the reported results.

%The discrepancy between the perceived performance and actual performance of neural counterpart of MVPS lies in the fact that accuracy in the DiLiGenT-MV~\cite{LiZWSDT20} benchmark is computed by measuring a bidirectional Hausdorff distance between the full surface of the predicted 3D shape and the full surface of the ground truth 3D shape, including the bottom of the object which is not visible in any frames. As one can see however in Figure~\ref{fig:intro} the average error is in fact dominated by the error in the bottom of the object and once discounted from the computation the classical method of~\cite{LiZWSDT20} achieves a better reconstruction error than PS-NeRF~\cite{psnerf}. Whereas our method obtains \textcolor{red}{0.2mm} presenting a new SOTA result for the multi-view photometric stereo.

In this work we provide the first neural multi-view photometric stereo approach which fully leverages the availability of pixel intensity information for estimating 3D shape from Photometric Stereo images (see Figure~\ref{fig:diligentqualitative}). We achieve this by explicitly modeling the incident light from point light sources to leverage intensity based shape optimisation over purely normal driven shape optimisation~\cite{psnerf,cao2023supernormal} which is fragile to incorrectly estimated normals especially  in cases of very few available lights as shown in Figure~\ref{fig:intro} and Section~\ref{sec:experiments}. %(especially for materials with complex BRDF function, e.g. metal or ceramics).

%We model point light attenuation, explicitly raytrace cast shadows and also optimise a fully neural material renderer. 
In more detail, we model point light attenuation and explicitly raytrace cast shadows in order to best approximate the incoming radiance for each point. The estimated incoming radiance is
used as input to a fully neural material renderer that uses minimal prior assumptions and it is jointly optimised with the surface. Estimated normals and segmentation maps are
also incorporated in order to maximise the surface accuracy. This allows us to achieve SOTA reconstruction accuracy (0.2mm) on original \textit{dense} (20 views, 96 lights) DiLiGenT-MV~\cite{LiZWSDT20} benchmark and signifcantly outperform (0.38mm vs 0.61mm) the best MVPS method~\cite{cao2023supernormal} in a \textit{sparse} setup of 6 views and 6 lights.% \textcolor{red}{Say something how learned structured renderer allows for extrapolating rendering in novel views out of the training trajectory Figure~\ref{fig:qualitative_brdf}}

%In addition, we  and use them to approximate the surface %\textit{ambient occlusion} and constrain self reflections. 

 %Our method consists of three steps: (1) estimating per-view shape from SOTA single view normal estimation network PX-Net~\cite{Logothetis22} (2) initialising a neural SDF shape with the shapes estimated during the first step, followed by (3) fitting the initialised neural SDF to image intensity and estimated normals. 
 
 %\textcolor{red}{The key difference of our method from the MVPS neural methods is [XX] for PS-NeRF~\cite{psnerf} and [YY] for OTHER]. It is similar to general NeRF~\cite{mildenhall2020nerf} such as NeILF++~\cite{zhang2023neilfpp} or [ZZ] however these methods do not correctly model light sources and learn them instead.} \\

%\noindent
%Our contributions are as follows:\\
%\begin{itemize}
%    \item{A novel neural multi-view photometric stereo method with correct point-light source modelling incorporating light attenuation and shadow computation.}
 %   \item{\textcolor{red}{Proposing an efficient neural caching procedure for estimating shadows txo speed up our training process by 2 times}.}
    %\item{Highlighting the poor protocol used to evaluate multi-view photometric stereo methods which includes the error computation on the inivisible part of the surface impacting the perception of the quality of competing algorithms and the difficulty of the benchmark.}
    %\item{Achieving the first SOTA results on the improved metric on the DiLiGenT-MV~\cite{LiZWSDT20} benchmark.\\}
%\end{itemize}

The remainder of this paper is organised as follows. Section~\ref{sec:relatedworks} discusses the related work in Photometric Stereo and Multi-View Stereo. It is followed by a description of our method in Section~\ref{sec:method}. The experimental setup and experiment results are described in Sections~\ref{sec:experimentsetup} and~\ref{sec:experiments} respectively.

%% file: sections/relatedworks.tex
%\newpage
\section{Related Work}
\label{sec:relatedworks}

There is an extensive literature on single and multi-view photometric stereo  and we review the following cases:%Some of these works deal with image-based models \cite{FurukawaP10,LhuillierQ05} where color texture at each pixel are projected depending on the retrieved depth.

\noindent
\textbf{Single view photometric stereo.} The first successful  deep learning based single view PS was CNN-PS   \cite{ikehata2018cnn} which was extended by \cite{logothetis2021pxnet} and \cite{Logothetis22} to be applicable to general calibrated point like configurations. Other works  like \cite{GoldmanCHS10,EstebanVC08,YakunLearning2023} have used material reflectance priors (using specific BRDFs like Lambertian or Ward) for single view normal prediction. Other recent approaches have leveraged the power of recent transformer models and big synthetic datasets (often of tens of thousand of images) to tackle a weakly uncalibrated setting like \cite{chen2019SDPS_Net,li2022selfps,li2022neural,yang2022snerf} and more recently fully uncalibrated single-view PS \cite{park2021nerfies, ikehata2023sdmunips}  and \cite{hardyunips}. However, despite the success of these methods in producing single view normal maps (as well as material maps), accurate shape
reconstruction is still challenging.
%Recentlyto PX-NET~ which is also extended for the near-field setting. 
%However, as textures have baked in light reflection, realistic re-rendering is hard to achieve. Furthermore, color texture relies on the material properties of the object, thus same image portion of the target object can appear substantially different from diverse views.
\noindent

\noindent
\textbf{Multi-view photometric stereo.} To overcome the ill-posedness of single view photometric surface reconstruction, multi-view photometric stereo (MVPS) methods have leveraged information from multiple view and multiple lights. Classical optimisation approaches have used triangle meshes \cite{park17mvps} or sign distance function based parameterisations \cite{Logothetis19,NiessnerZIS13,ZollhoferDIWSTN15} to tackle the multi-view PS problem, under diffuse reflectance. Methods, e.g. \cite{LiZWSDT20}, were also developed for more general materials as well.

\noindent
\textbf{Neural surfaces.}  Recently, neural surface approaches have became very popular in tacking the 3D reconstruction problem.  Early approaches include NeRF \cite{mildenhall2020nerf} and its first extensions to neural SDF parameterisations~\cite{yariv2020multiview, yariv2021volume,wang2021neus}. The first methods which used neural SDFs specifically for the multi-view PS problem include~\cite{Kaya2021,kaya2022uncertainty,kaya2023multi,Zhao_2023_ICCV}. However, contrary to the direction of the neural inverse rendering literature, these approaches do not attempt to re-render the original photometric stereo images but rather some 2D derivates of the images such as normals or albedo maps. For example, Supernormal \cite{cao2023supernormal} only renders normal maps but achieves very fast training speed though patch parameterisation, as well as the use of the NERFACC~\cite{li2023nerfacc} framework. PS-NeRF \cite{psnerf} renders normal and average intensity maps whereas RNb-Neus \cite{BrumentRNb24}, uses normal and albedo maps to render virtual light images. Thus, all these methods are reliant on single view PS networks and have no way to circumvent noisy estimates that are likely to happen in case of sparse lights and number of views as demonstrated in Figure~\ref{fig:intro}. %such as normal maps, average intensity maps albedo maps. 
 %and SuperNormal \cite{cao2023supernormal} .
 %However, these approaches are mainly relying on normal fusion and do not attempt to update the neural surface though rendering.

Other recent neural rendering approaches have advanced the sophistication of the rendering methods to be more structured and thus respect the physics of light reflection more, like Ref-NeRF~\cite{refnerf}, Neuralangelo~\cite{li2023neuralangelo}, NERO~\cite{liu2023nero} and NeILF++~\cite{zhang2023neilfpp} but none of these methods has yet to be applied to PS problem, especially lacking the ability to model point light illumination.  Finally it is worth mentioning \cite{guo2022edgepreserving}  who introduced the idea of a infinitely differentiable surface (SIREN \cite{sitzmann2019siren}) with Lambertian rendering for the single view PS and \cite{Logothetis24WACV}, which extended this method to the binocular setting and also added a fully learnable general material renderer.  

We borrow the material renderer from \cite{Logothetis24WACV} while extending the approach to work in the multi-view setting using an SDF parametrisation similar to \cite{yariv2020multiview, yariv2021volume}. It is worth noting that Brahimi et. al, \cite{Brahimi_2024_CVPR}, also tackles the MVPS problem though physics-based per point rendering (using the Disney BRDF~\cite{burley2012physically}). This approach is the most similar to us with major differences being that we use a fully neural material model, we explicitly ray-trace cast shadows and also employ supervision signal from single view photometric stereo normals. Thus we are able to outperform them with 0.25mm vs 0.34mm error (see Table~\ref{tab:Tab_eval_diligent}).  %SLU Sparse semi-calibrated lights

%\noindent
%\textbf{New ones.}

%Volume Rendering of Neural Implicit Surfaces, original idea for sdf for surfaces

%This is the official implementation of Neuralangelo: High-Fidelity Neural Surface Reconstruction

%NeRO: Neural Geometry and BRDF Reconstruction of Reflective Objects from Multiview Images .
%MVPSNet: Fast Generalizable Multi-view Photometric Stereo bad and retarded  

% NeILF++ is a differentiable rendering framework for joint geometry, material and lighting estimation from multi-view images %https://github.com/apple/ml-neilfpp

%Introduced a neural-heihtmap approach with near-field point light rendering and a fully learnable general BRDF parameterisation 

%Nvidia neural caching \cite{muller2021real}

%yvain arxiv new good \hyperlink{https://arxiv.org/pdf/2312.01215.pdf}{https://arxiv.org/pdf/2312.01215.pdf}

%n%erf -- calibration  \hyperlink{https://arxiv.org/pdf/2102.07064.pdf}{https://arxiv.org/pdf/2102.07064.pdf}

%https://openaccess.thecvf.com/content/CVPR2023/papers/Cao_Multi-View_Azimuth_Stereo_via_Tangent_Space_Consistency_CVPR_2023_paper.pdf

%super normals ,better version of psnerf, very fast https://arxiv.org/pdf/2312.04803.pdf

%roboticsapplication https://openaccess.thecvf.com/content/CVPR2023/papers/Cheng_WildLight_In-the-Wild_Inverse_Rendering_With_a_Flashlight_CVPR_2023_paper.pdf

%yvain bug transformer https://tschumperle.users.greyc.fr/publications/tschumperle_hal24.pdf 
%https://github.com/Clement-Hardy/Uni-MS-PS/tree/main 

%% file: sections/method.tex
\section{Method}
\label{sec:method}

\begin{figure*}[t]
\includegraphics[width=\textwidth]{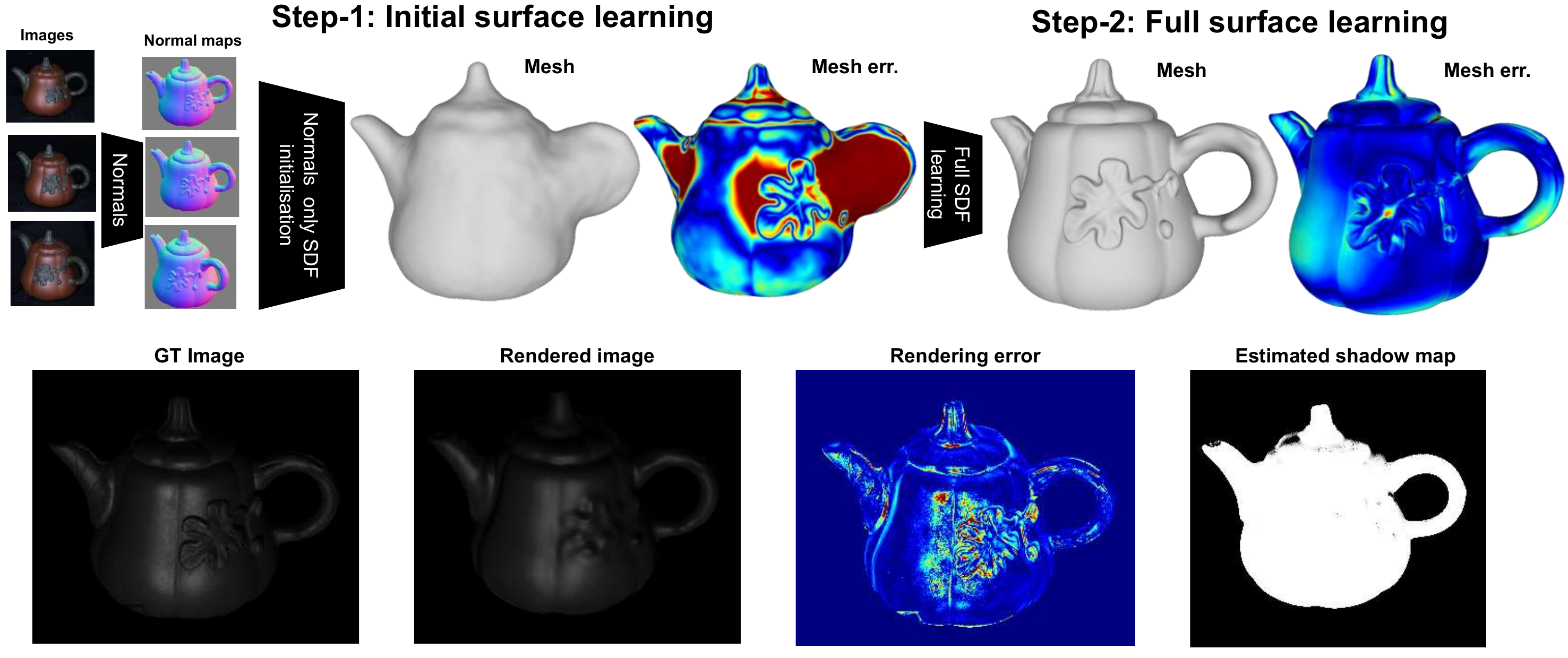}
\caption{Schematic of our overall method. Single view PS is used to obtain normal maps. Training the SDF with normal and silhouette loss (for 3 epochs only, see Section~\ref{sec:init}) obtains a rough surface which is then refined with full volumetric rendering, explained in Figure~\ref{fig:diagram}. The second row also shows the GT and render images (as grayscale), the rendering error (with red $\ge 0.1$) as well as the computed shadow map.}
\label{fig:method}
\end{figure*}

This section describes our method for solving the point light multi-view photometric stereo. A high level overview is also shown in Figure~\ref{fig:method}.  Our method is primarily an inverse neural rendering method. Section~\ref{sec:irad} describes the assumed irradiance model. Sections~\ref{sec:sdf} and~\ref{sec:init} describe the underlying neural surface parametrisation and its initialisation, respectively. Section~\ref{sec:losses} describes training losses used.

%\noindent
\subsection{Irradiance equation}
\label{sec:irad}

We now explain the assumed irradiance equation of a world point $\vect{X}$ with surface normal $\vect{N}$ and albedo $\rho$. We assume point light sources $m$ at positions $\mathbf{P}_m$  which generate variable lighting vectors $\mathbf{L}_m=\mathbf{P}_m-\mathbf{X}$. In addition, point light propagation results to the following attenuation factor $a_m=\frac{\phi _m  }{||\mathbf{L}_m||^2}$ where $\phi _m$ is the intrinsic brightness of the light source. We note that the literature~\cite{Mecca2014near} usually also assumes angular dissipation factor but these calibration numbers are unavailable for DiLiGenT-MV \cite{LiZWSDT20} therefore we opt for a simpler, perfect point light source model. Thus, the reflected intensity of the point $\vect{X}$ % (not include $\vect{X}$ for clarity) 
for the  $m$-th light source $i_m$ is modelled as:

\begin{equation}
i _m=s_m a_m \rho B (\vect{N},\vect{L}_m,\vect{V}) 
\label{eq:irad}
\end{equation}

Note, here $B(.)$ is assumed to be a general BRDF, $s_m \in \{0,1\}$ is an indicator variable to account for cast shadows that completely block direct reflectance. We assume that indirect reflectance (i.e. self reflections and ambient light) are negligible and can be ignored. Also note that analytic BRDF models like \cite{CookTorrance1982,burley2012physically} often (but not always \cite{mecca2016single}) separate diffuse and specular components and may include separate albedos. However, a completely learned BRDF as proposed in Section~\ref{sec:sdf} does not need to follow this structure.

%$i_{r,m}$ corresponds to indirect reflectance (e.g. self reflections) and we will further make the simplifying assumption that it is also proportional to the surface albedo as well as the light source attenuation.  
%\textcolor{red}{no direct reflectance-no rend loss shadow}
%Finally, we note that self reflection occur on concave parts of an objects surface which also correlate with more cast shadows therefore we make the following  assumption  $i_r \propto (a \rho \sum (1-s))$. Indeed, the term \textit{ambient occlusion} is usually used to describe the total obliqness of a surface point and this is approximated with the number of shadows. To simplify the learning problem, we assume that $i_r$ is constant between different lights thus  $i_{r,m}= a \rho \sum s R (\vect{X})$. This assumption is further justified empirically on the supplementary by empirical measurements on Blender-rendered data.

\subsection{Neural SDF}
\label{sec:sdf}

\begin{figure}[h]
\includegraphics[width=\columnwidth]{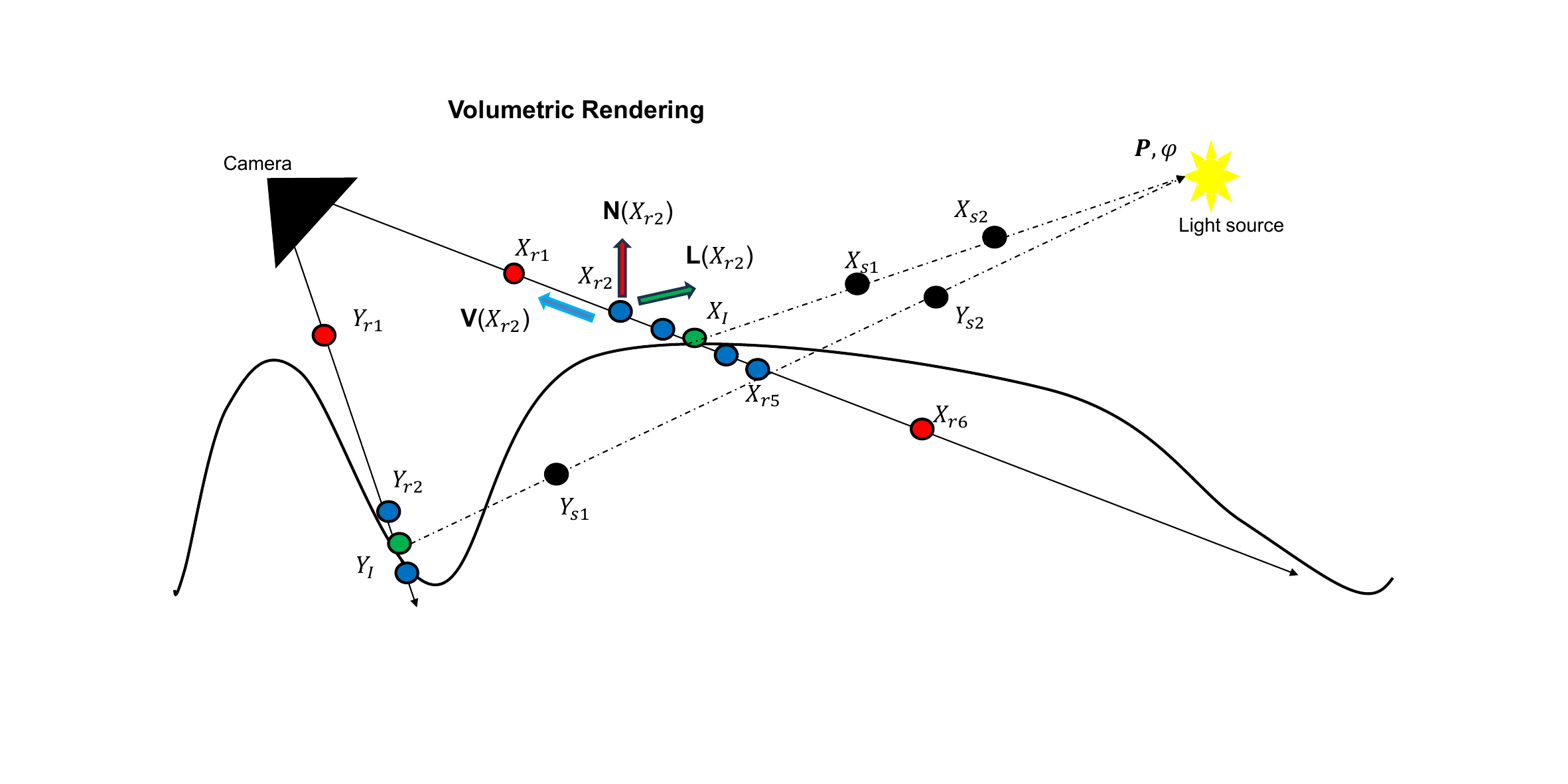} \vspace{-1.5cm}
\caption{Visualisation of our volume rendering approach. Two rays with multiple ray samples $\vect{X}_{ri}$, and $\vect{Y}_{ri}$ are shown. The surface-ray intersection points $\vect{X}_I$, and $\vect{Y}_I$ are also shown as they are used to ray trace cast shadows (towards the light source at position $\vect{P}$ with brightness $\phi$). Cast shadow samples are marked as  $\vect{X}_{si}$, and $\vect{Y}_{si}$ respectively. Note that points that significantly contribute to the total rendering (though the accumulated opacity) are coloured blue and points that do not (because they are outside of the surface or occluded) are marked red. For shadow sample points rendering is not performed and so are marked black. Note that the intersection points ($\vect{X}_I$, and $\vect{Y}_I$) are only used to guide shadows so they are not rendered either. Finally, for the  $\vect{X}_{r2}$ ray sample point, normal $\vect{N}$, lighting $\vect{L}$ and viewing vectors $\vect{V}$ (that are used for rendering) are shown with respective colors of red,green and blue.}
\label{fig:diagram}
\end{figure}

\noindent
\textbf{Geometry parameterisation.} Following the work of other neural volumetric approaches \cite{yariv2020multiview,yariv2021volume}, we parameterise the scene geometry as the zeroth level set of an implicit function $F$ corresponding to the Signed Distance Field,  $d = F (\vect{X})$, %. The unknown function $F$ is 
parameterised by a deep neural network. % and the objective is to optimise its weights.   
We note that the SDF of any arbitrary geometry is always continuous, almost everywhere differentiable and satisfies the Eikonal equation of unit magnitude gradient $|| \nabla F(\vect{X}) ||= 1$. Finally, we note that for surface points (where $ F (\vect{X})=0$), the surface normal is the gradient of the SDF, i.e.  $\vect{N}(\vect{X}) =  \nabla F $. This allows to train the SDF through rendering loss from the initial photometric stereo images. In addition, if surface normal maps are available (e.g. from single view estimation networks) they can also be used as an addition training signal.  We use the SIREN architecture \cite{sitzmann2019siren} which is a MLP with sinusoidal activation functions and that guarantees that the surface is infinitely differentiable thus can be easily recovered from its derivatives. %Indeed,  \cite{guo2022edgepreserving,Logothetis24WACV,zhang2023neilfpp} have used SIREN networks for single and binocular view PS as well as multi-view stereo so we are a direct extension to them specifically for the MV-PS under point light sources.

\noindent
\textbf{Ray sampling}. We follow a volumetric sampling and rendering method similar to VolSDF~\cite{yariv2021volume} where the neural SDF is queried in multiple samples on outgoing rays from each image foreground pixel. For each pixel an estimate of the depth is available (initialised from single view PS and occasionally updated during training) and thus most of the samples are concentrated around that depth. However, to allow the surface to evolve and to minimise free space artefacts, additional samples are also sampled in a wider depth range. Following VolSDF~\cite{yariv2021volume}, we use the Laplace density function to convert from SDF values $d$ to density $t$ as follows:
\begin{equation}t(d)=\big(0.5 + 0.5 \cdot sign(d)  (\exp(-|\beta d|)-1)\big)/\beta \end{equation} with $\beta$ being a trainable scalar constant controlling the sharpness of the distribution. 

Finally, alpha blending is used to accumulate depth, normals and rendered intensity at each ray using the standard approach, i.e. transparency $\alpha = \text{exp} (-t \delta r) $ with $ \delta r$ being the distance along the ray and so the surface-ray intersection point $\vect{X}_I$ is computed as a weighted sum of ray samples $\vect{X}_I= \sum_{i} (w_i \vect{X_r}_i )$, with the sample weights $w_i$ corresponding to the accumulated opacity. Note that intersection points $\vect{X}_I$ are further used to compute cast shadows but are not directly rendered. Similar averaging is used to obtain the rendered intensities  ($i_I= \sum_{i} (w_i i_i )$) and surface normals ($\vect{N}_I= \sum_{i} (w_i \vect{N_r}_i )$) both of which are used to compute losses.  The ray sampling process is further explained in Figure~\ref{fig:diagram}.

%\textcolor{red}{intenisty and depth, explain we need this average point for shadows and depth estimate} %The surface normal is computed in  a similar manner to the position $\vect{p}$  and can be projected back to camera coordinates with the transpose rotation $R^\intercal \cdot\vect{n}_e $.    
 
\noindent
\textbf{Learned BRDF}. We follow the approach of \cite{Logothetis24WACV} where the BRDF is also parameterised as another SIREN network and thus is completely learned from the data. This assumes that the material properties are uniform around the target scene except for a scalar albedo variation.  We emphasise that we chose to perform grayscale intensity rendering instead of full RGB rendering as this is expected to minimise the synthetic to real gap. Real RGB images are usually acquired with demosaicing of single intensity values and this procedure is usually optimised to best recover intensity not colour (e.g see \cite{getreuer2011malvar} used by OpenCV).

\begin{figure}[t]
\includegraphics[width=\columnwidth]{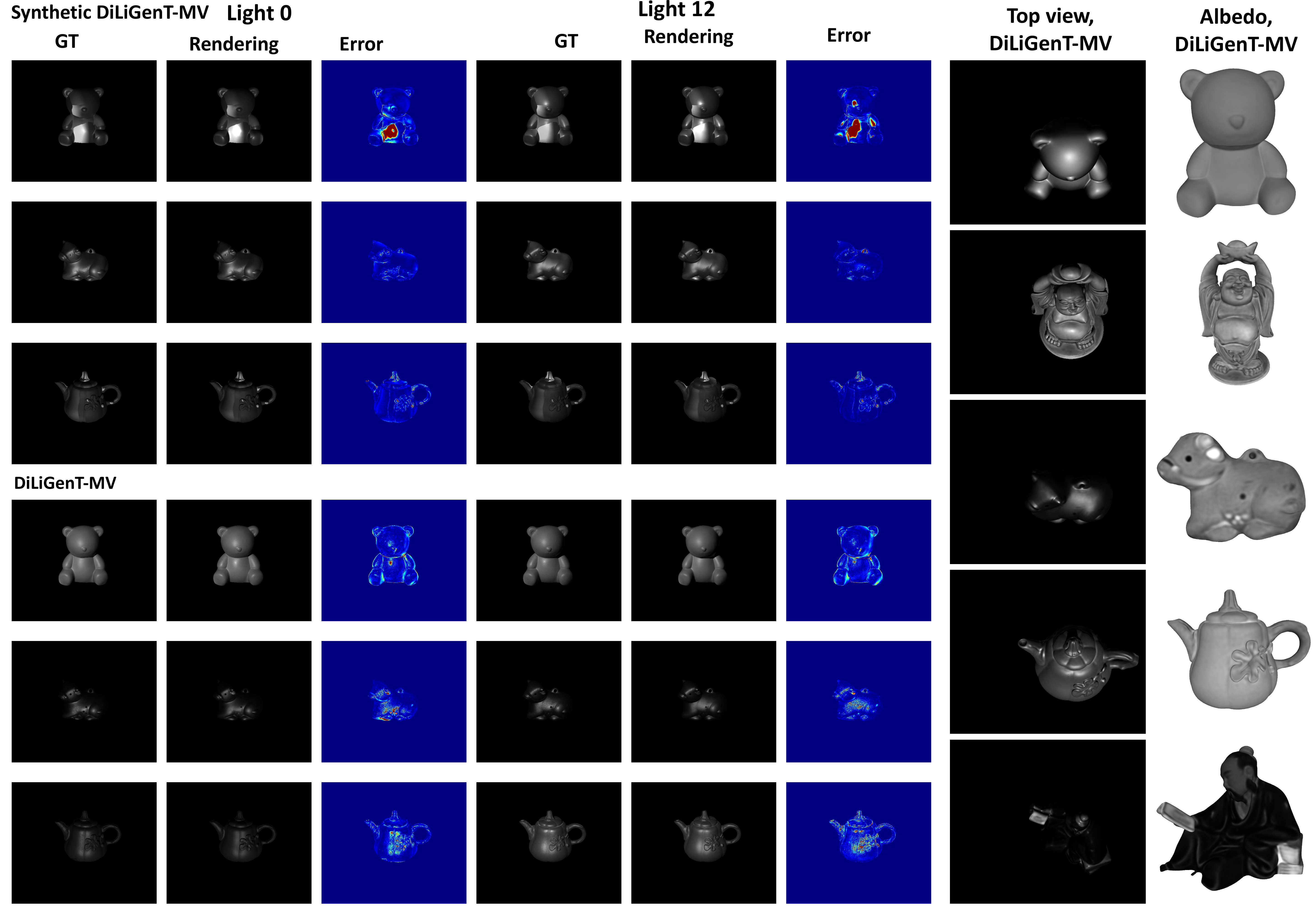}
 \caption{Qualitative visualisation of re-rendering and rendering errors for synthetic and real data (left side, top and bottom 3 rows). The scaling of the error map sets red to $\ge0.1$. We note that most of the error is concentrated on the middle of concavities as self reflection is not modeled. On the right side we see renderings in a novel angle of objects recovered from real data as well as recovered albedo maps.}
\label{fig:qualitative_brdf}
\end{figure}

To minimise over-fitting, the material BRDF network receives as input only the relative angles between $\vect{N}$, $\vect{L}$ and $\vect{V}$. In addition, to simplify the learning problem, we follow the principles described in the MERL database \cite{Matusik2003jul}. To achieve this, the half vector $\vect{H}=\frac{\vect{L}+\vect{V}} {|\vect{L}+\vect{V}|} $ is first computed and the input to the network is the relative angles between $\vect{N}$, $\vect{L}$ and $\vect{H}$. Finally, we note that the final activation of SIREN part of the BRDF network is exponential and there is a post multiplication with an $\vect{N} \cdot \vect{L}$ factor so that the BRDF network learns a multiplicative factor over the diffuse reflectance, parameterised as follows:
 \begin{equation}
\text{B} (\vect{N},\vect{L}_m,\vect{V}) =  (\vect{N}\cdot \vect{L}_m) \text{ exp\big{(}SIREN} ( \vect{N}\cdot  \vect{H}_m, \vect{N}\cdot  \vect{L}_m, \vect{H}_m\cdot  \vect{L}_m) \big{)}
\label{eq:brdf}
\end{equation}
%\sphericalangle
\textbf{Albedo.} The scalar albedo $\rho$ is learned with another SIREN network which is queried for every sample point. We note that having the BRDF network constant throughout the volume and only varying a scalar albedo may sacrifice quality in objects with significantly varied materials, but this does not seem to be the case in the DiLiGenT-MV~\cite{LiZWSDT20} as shown in Figure~\ref{fig:qualitative_brdf}. We note some competitors like Neuralangelo~\cite{li2023neuralangelo} learn a fully-varied rendering network parameterised by  position, normal, lighting and viewing vectors, but this approach is a lot more prone to over-fitting and would struggle to extrapolate the rendering into completely unseen viewing angles, which is not the case for our approach (see Figure~\ref{fig:qualitative_brdf}).
%\begin{figure}[h]
%\%includegraphics[width=\columnwidth,height=5cm]{training_dynamics.png}
%\caption{ROC figure for training dynamics/time - with caching vs without cashing}
%\end{figure}
 
 %The subscript $w$ is used to indicate world coordinate frame as opposed to $c$ for camera coordinates; converting between coordinates with rotation $R_c$ and translation $\vect{t}_c$

%\begin{equation}
%[x_w,y_w,z_w]^\intercal = R_c \cdot [x_c,y_c,z_c]^\intercal + \vect{t}_c
%\label{eq:coordinate}
%\end{equation}

\textbf{Shadow estimation.} %Handling cast shadows is imperative for any successful photometric stereo method. 
To estimate cast shadows for a ray-surface intersection point $\vect{X}_I$, we raytrace from that point to the light source following the direction of the lighting vectors $\vect{L}_m$ computed above.  For each ray we take 16 random samples $j$ $\in [2\text{mm}, 50\text{mm}]$. For all these points, we query the SDF network and accumulate opacity following the same volumetric rendering procedure, i.e $s = \prod_j (1-\alpha_j)$. We note that this shadow computation procedure is very computationally expensive therefore it cannot be computed for all the rendered points. Instead, we only compute it for the intersection point of each ray ($\vect{X}_I$ is Figure~\ref{fig:diagram}) and assume it is the same for all other ray samples. 

%\textbf{Indirect reflectance.} Finally, the indirect reflectance $i_{r,m}$ (Equation~\ref{eq:irad}) is approximated as  $i_{r,m}= a \rho \tilde{s} R (\vect{X})$. $\tilde{s}$ is compute as the mean value of one minus shadows (over the number of lights) and its a measure of the local surface concavity; $\tilde{s}=0$ corresponds to no shadows and hence completely convex local geometry therefore no intereflection,  $\tilde{s}=1$ corresponds to all lights being shaded and hence maximum concavity therefore we allow for maximum learned intereflection. Finally $R (\vect{X})$ is queried from another SIREN network and we note that it is parameterised to be constant on all lights. 
Visualisation of the generated renderings of synthetic and real data are shown in Figure~\ref{fig:qualitative_brdf}.

 %Extending  \cite{guo2022edgepreserving} to the 2 view problem, we chose the SIREN architecture \cite{sitzmann2019siren} which is a MLP with sinusoidal activation functions and that guarantees that the surface is infinitely differentiable thus can be easily recovered from its derivatives; thus the surface normal isand automatic differentiation makes it a function of the network weights. We also add an additional scalar (grayscale) albedo $\rho= F(x_s,y_s)$ channel on the SIREN used for rendering.

\subsection{Initialisation}
\label{sec:init}

As it is standard in MVPS approaches, e.g.~\cite{psnerf}, we can use single view PS (at each view) in order to obtain normal and depth estimates. We start by computing per view normal maps using the state-of-the-art PS normal estimation network~\cite{hardyunips} and also use numerical integration \cite{queau2015edge} in order to get approximate depth maps. The normal maps are used in order to provide an additional training signal. 

The depth maps do need to be accurate as they are only used to initialise the ray sampling space. More specifically, classical ray marching (e.g. NeRF \cite{mildenhall2020nerf}) uses fixed near/far planes for each pixel which is inefficient while newer approaches use an occupancy grid (e.g NERFACC~\cite{li2023nerfacc}) to guide the search space. We opt for a simpler solution where for each pixel, near/far planes are centered around the pixels depth estimate; this is initialised with single view depth estimate and updated every 5 epochs.

The SDF network is initialised with weights that approximate the SDF of a perfect sphere. To speed up convergence, we always run 3 epochs with normal and silhouette loss without rendering. 
%This method offers a general near-field network that obtains high quality normal maps for the calibrated, near (and far) field PS setting as well as some reasonable surface estimate. %See example for estimated normals in Fig.~\ref{fig:method} and supplementary material.

\noindent
\textbf{Final surface calculation.} After the optimisation is completed, the SDF network can be sampled in a regular grid of points and a triangle mesh surface can be recovered using the standard Marching Cubes \cite{lorensen1998marching} algorithm. For recovered surface points, the albedo network is queried in order to obtained a textured reconstruction. %these revered vetrices, the albedo network can be queried in order to obtained a textured reconstruction.%  \textcolor{red}{fdo we need that?}

%\subsection{Training}
%\\label{sec:training}

\subsection{Losses}\label{sec:losses}

We use the following losses.

\noindent
\textbf{Rendering loss.} We use L1 loss on the rendered intensities, i.e. $L_{rend}=|i_{rend} - i_{gt}| $. We note that to better balance the rendering data, each light source is scaled so as the maximum GT intensity is 1 (saturated). This is performed because some of the DiliGenT-MV lights are very dim. The relative weighting of this loss when used is 1000.

\noindent
\textbf{Silhouette loss.} Similar to previous approaches \cite{wang2021neus,li2023neuralangelo}, we also apply binary cross entropy loss,  $L_{sil}=\text{BCE}( \alpha _{total}, mask) $, between the predicted silhuettes and the ground truth ones. We generate predicted silhuettes with the total accumulated opacity $\alpha _{total} = \prod_j (1-\alpha_j) $ (which should be 1 for foreground and 0 for background points),  We consider a 10 pixel band outside the provided segmentation masks for background points. The relative weighting of this loss is 10. %We note that we ignore the utmost pixels of the provided segmentation masks and also the innermost pixels of the background (leaving 2 pixel boundaries between foreground and background). This is done for better numerical stability as the exact discritisation of the boundaries is not very reliable.  

\noindent
\textbf{Normal loss.} We apply angular normal loss to match SDF computed normals $\vect{N}_s$ to the network predicted normals  $\vect{N}_n$. We follow \cite{Logothetis24WACV, logothetis2021pxnet} and use angular error (as opposed to L2 loss used in PS-NeRF~\cite{psnerf}) $L_{n}=|\text{atan2}(||\vect{N}_n \times \vect{N}_s||, \vect{N}_n \cdot \vect{N}_s)|$.  For rays where the accumulated opacity is less than 0.01 (i.e the ray does not intersect any surface), no normal loss is applied. In addition, following previous works, the normal loss is weighted by the obliqueness of each point  $(\vect{N}\cdot \vect{V})$ and that stops the optimisation to try to fit occlusion boundaries which are numerically unstable. The relative weighting of this loss when used is 1.

\noindent
\textbf{Eikonal loss.} To enforce the Eikonal equation for all ray samples, we apply L1 loss which is the standard in most SDF approaches i.e. $L_{eik}=| ~|| \nabla d ||- 1| $. The relative weighting of this loss is 10.

\noindent
\textbf{Curvature regulariser.} To minimise floater artefacts (especially on the inside of objects) and encourage the optimisation to recover the minimum surface, it important to include some curvature regulariser encourage smoothness on the volumetric normals (i.e. SDF gradients). Computing exact analytic curvatures (via auto-differentiation) has a high computational cost an is not really required, as the objective is to only use them as a regulariser. Instead, inspired by SuperNormal~\cite{cao2023supernormal}, we use the ray samples $\vect{X}_i$ and compute finite differences along the ray as: $curv(\vect{X}_i)\approx \frac{|| \vect{N}_i - \vect{N}_{i+1}||}{||\vect{X}_i - \vect{X}_{i+1} ||}$. We note that an exact curvature would require finite differences along all 3-axis, but for regularisation purposes this definition is adequate, and comes with no additional SIREN queries. Note that NeILF++~\cite{zhang2023neilfpp} and Neuralangelo~\cite{li2023neuralangelo} also include a similar regulariser.  The relative weighting of this loss is 1.

%% file: sections/experimentsetup.tex
\section{Experiment Setup}
\label{sec:experimentsetup}

\begin{table}[t]
\setlength{\tabcolsep}{3.0pt} 
\begin{center}
\begin{small}  %footnotesize
\begin{tabular}{||c| c c c c c ||c|| }%c|c||} 
 \hline
 Method & Bear & Buddha & Cow & Pot2 & Reading & Avg. SE \\ %& Avg. A-ZE & Med. A-ZE  \\ 
 \hline\hline
  \hline
 
  GT Normals  & 0.06	&  0.08 & 	0.03 & 0.04	&  0.02	&  0.05	 \\ % &  &    \\ 
\hline
N  & 0.17	& 0.14	& 0.10	& 0.14	& 0.16	& 0.14 \\
%I - shd. & -	& -	&-	& -	& -	& -\\
I &  0.13	& 0.23 &	0.06 &	0.08 &	0.12 &	0.12\\ 
I-S & 0.20	& 0.27	& 0.04	& 0.08	& 0.18	& 0.15 \\
N + I  & 0.11	& 0.15	& 0.03	& 0.07	& 0.11	& 0.09\\

 \hline

\end{tabular}
    \vspace{-0.1cm}
    \caption{ Ablation study of our method on a synthetic replica of DiLiGenT-MV~\cite{LiZWSDT20} benchmark.  We first compute our method performance using ground truth normals in order to highlight potential issues with real DiLiGenT-MV~\cite{LiZWSDT20} benchmark (first two rows of Table~\ref{tab:Tab_eval_diligent}) where recovered shape is significantly less accurate both for our and Supernormal~\cite{cao2023supernormal} methods (0.05mm vs 0.11mm). 
    We also include comparison of four versions of our method named (N), (I), (I-S) and (N+I). %namely normal loss only (N) - no rendering, rendering loss only (I) - no normals loss, rendering with no shadows I-S (and no normals) and combined (N+I).
    (N) corresponds to only applying normals loss, where (I) and (I-S) corrspond to only using rendering loss with and without shadows respectively. (N+I) combines all losses. The combined approach achieves the best error and particularly note that it outperforms both other configurations most objects indicating the the combined approach is better than a simple interpolation between the two.
}
    %Using normals and intensity rendering for shape estimation achieves the best error in both \textit{sparse} and \textit{dense} setups, respectively 0.21mm and 0.12mm. %Note jointly using normals and intensities outperforms other configurations on Bear and Cow objects indicating the the combined solution is better than a simple interpolation between the two.  
    \label{tab:Tab_eval_syntheticdiligent}
\end{small}
\end{center}
\end{table}

%This section explains the datasets used for evaluation of our method along with the evaluation metric and competitors. 
This section describes the datasets as well as the training and evaluation protocol.%metrics, and competitors used to assess our method.

\subsection{Datasets}\label{sec_datasets}

\noindent
\textbf{DiLiGenT-MV.} Our main evaluation is performed on DiLiGenT-MV \cite{LiZWSDT20} benchmark containing 5 objects with 96 lights in 20 views. Images are of $612\times512$ px resolution with objects actually occupying a maximum of $400\times400\text{px}$. Ground truth meshes, camera intrinsics, extrinsics and normal maps are provided, together with point light positions and far-field light brightnesses ($\phi$). We note that these brightnesses were measured from the intensity of a flat calibration target roughly positioned at the location of the imaged objects, so intrinsic brightness is recovered by multiplying with inverse distance square. We note that as such calibration data is unavailable, the $\phi$ is expected to be fairly inaccurate and thus it is optimised during training. 

\noindent
\textbf{Dataset anomalies.} In order to have the most fair assessment, we report the following dataset anomalies on DiLiGenT-MV  and our attempts to overcome them. 

Firstly, the provided GT normal maps and masks are incompatible with renderings of ground truth meshes when provided intrinsic and extrinsic parameters are used. %The best compatibility was achieved by using the intrinsic matrix for Reading for all objects while simultaneously performing SDV to fix the non-one determinant on the provided rotation matrices (especially for Reading).
Note, some of the provided rotation matrices are not orthonormal and have non-unitary determinant.  We follow the approach of RNB~ \cite{BrumentRNb24} and Supernormal~\cite{cao2023supernormal} by first performing an explicit projection matrix computation ($P=K[R|T]$) followed by QR decomposition (using OpenCV) to obtain orthonormal rotation matrix. % Despite that, there is still some mismatch between provided normals, masks and meshes, thus potentially limiting the maximum achievable accuracy.
%For all experiments expect the first line of Table~\ref{tab:Tab_eval_diligent}, we use the modified parameters (Reading instrisics and endorsing unit determinant on rotation matrices R with SVD).

Secondly, the provided segmentation masks in Bear and Cow contain holes that need to be closed manually in order to prevent the silhouette loss from introducing large holes in the reconstructed meshes. 

Finally, Ikehata et al~\cite{ikehata2018cnn} first noticed that the first 20 images of the Bear appear to be corrupted. We also found more similarly corrupted images on other views (more visualisations in the supplementary) and did our best effort to manually mark and ignore them however it is possible that more image corruptions are still unnoticed.

\noindent
\textbf{Synthetic DiLiGenT-MV.} To better demonstrate the effectiveness of our method without the real data corruptions discussed above, we rendered a synthetic version of DiLiGenT-MV with Blender. See Figure~\ref{fig:qualitative_brdf}. We use the exact same objects, with the exact same poses and rendered the 96 points lighst. The objects materials where chosen to loosely mimic real objects and the albedo was set to a random pattern. Finally, we note that this synthetic data can be used to visualise shadow and indirect reflection maps which are really hard to correctly evaluate on real data. %This is explored in the supplementary.

\subsection{Hyperparameters and Training}

We use a tensorflow port of SIREN for all the experiments. The SDF MLP is set to 5x512 layers (1.05M parameters) and the albedo   MLP to 3x256 (133K parameters). The BRDF MLP is set to 3x32 layers (2.5K parameters). We use 64 ray samples in a 100mm ray range and an additional 64 around the average intersection (in a shrinking distance range up to 10mm) and 64 extra samples computed with one step of Newton method (for approximating the 0 of the SDF). For each shadow ray we used 16 samples. We train with batch size of 512 rays for 100 epochs which takes approx. 20h on a NVIDIA TITAN RTX and 17GB of RAM when rendering all 96 lights. Note that the 6 lights 6 views version completes in only 2h.

\subsection{Evaluation protocol}

We evaluate our method by computed Chamfer distance (marked as surface error SE) of the reconstructions and the ground truth. This is computed as the average of asymmetric Hausdorf distance from reconstruction to ground truth and the opposite, with the distances  computed with with Meshlab. We note that in order to have a fair comparison and not bias the error with unseen bottom of the objects, the bottom 6mm of GTs and all reconstructions are removed. All DiLiGenT-MV objects are aligned to be touching the XY plane so the cropping is straightforward. This cropping also avoids large error at some parts of the bottom of the objects that are occluded by the background (e.g. the feet of Reading, see Supplementary Material). Thus our reported error (in Table~\ref{tab:Tab_eval_diligent}) is generally \textit{lower} than the numbers originally reported in other works.  

\noindent
\textbf{Competing approaches.} We compare against DiLiGenT-MV~\cite{LiZWSDT20},   PS-NeRF~\cite{psnerf}, MVAS~\cite{cao2023mvas} Brahimi et. al, \cite{Brahimi_2024_CVPR} RNB~ \cite{BrumentRNb24} and Supernormal~\cite{cao2023supernormal}. DiLiGenT-MV~\cite{LiZWSDT20} and  Brahimi et. al, \cite{Brahimi_2024_CVPR} are closed source so we use the meshes computed by the original authors; for all other methods we use the meshes reproduced from their original codes. We note that Supernormal~\cite{cao2023supernormal} offers the best perfomance and by far the least computational cost so it is used for ablating the sparse lights and views scenario as well. Also Brahimi et. al, \cite{Brahimi_2024_CVPR} was only computed on the sparse scenario therefore that is the only available comparison for them. 

%For all of the methods, the output reconstructions were downloaded from the SuperNormal project page

%\footnote{\href{https://xucao-42.github.io/homepage/}{https://xucao-42.github.io/homepage/}}.

%Synthetic data is only used to ablate our own method as the DiLiGenT-MV and Supernormal code is not available (and our method using only norma loss is very similar to PS-NeRF and Supernormal). We also note that  PS-NeRF and MVAS are also using a modified/pre-procesed version of DiLiGenT-MV and this is potentially alleviating some of the anomalies (especially the segmentation mask holes). 

%use full light normals of other
%we only go up to 30

%% file: sections/experiments.tex
\section{Experiments}
\label{sec:experiments}

\begin{figure*}[t]
\includegraphics[width=\textwidth]{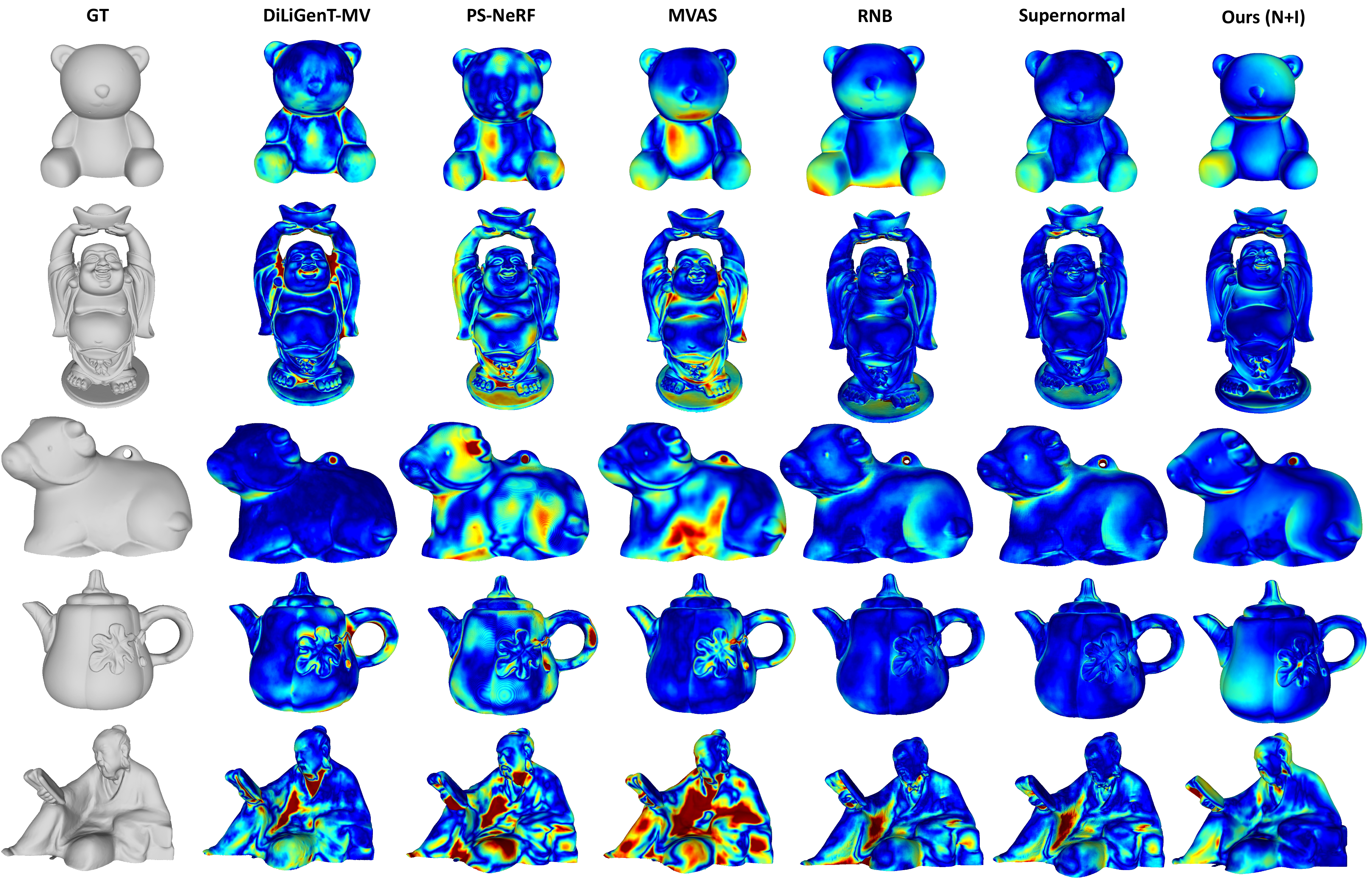}
\caption{Qualitative results on real DiLiGenT-MV~\cite{LiZWSDT20} benchmark. For each mesh vertex, the minimum distance to the GT mesh is shown with the error bars set to red corresponding to 1mm. We note that we are achieving consistent, uniform accuracy on all regions of all objects, including the concavity in the middle of Reading. }
\label{fig:diligentqualitative}
\end{figure*}

We describe two sets of experiments on synthetic data and real data in Sections~\ref{sec_results_artificial} and~\ref{sec_results_real} correspondingly.

\subsection{Synthetic data}\label{sec_results_artificial}

As mentioned in Section~\ref{sec_datasets}  the original DiLiGenT-MV~\cite{LiZWSDT20} dataset contains several anomalies making hard to correctly ablate the several steps of our method. %\textcolor{red}{has issues with its intrinsic and extrinsic parameters} so to prove that our approach works correctly we ran an ablation study on 
Thus additional ablation experiments are performed on Synthetic-DiLiGenT-MV dataset. % described in Section~\ref{sec:experimentsetup}.

We first show that our network can achieve a very low error of 0.05mm when using the ground truth normals which is not the case for the real DiLiGenT-MV dataset as shown in Table~\ref{tab:Tab_eval_diligent}. We also show that if predicted normals are used the performance is worse: 0.14mm. Using the rendering of intensities only achieves overall error of 0.12mm. In addition, using the rendering loss on intensities only (I) significantly outperforms the normals loss only (N)  experiments on all objects except Buddha. The reason for such a differing performance is that the presence of strong self reflection effects presents a more difficult task to the rendering network than for the normal estimation network. Note that not computing the shadow maps while using rendering only loss (I-S) increases the error from 0.12mm to 0.15mm, with the very concave object Reading being affected the most (0.12m to 0.18mm). The combined normals and intensity rendering experiment achieves the best accuracy of 0.09mm average error.  %In addition, on Bear and Cow objects, the accuracy using both losses is better than any of the individual ones (0.17mm, 0.13mm vs 0.11mm for Bear) and (0.10mm, 0.07mm vs 0.04mm for Cow) therefore showing the ability to combine best of both training signals.

%showing the ability to improve on the objects %(e.g. \textcolor{red}{aa, bb}) \textcolor{red}{where only normal estimation network was wrong}. The rendering visualisation of this network can be found in Figure~\ref{fig:qualitative_brdf} (top). \textcolor{red}{add more points regarding rerendering}

\begin{table}[t]
\setlength{\tabcolsep}{2.0pt} %3pt
\begin{center}
\begin{small}  %footnotesize
\begin{tabular}{||c| c c c c c ||c|| }%c|c||} 
 \hline
 Method & Bear & Buddha & Cow & Pot2 & Reading & Avg. \\ %& Avg. A-ZE & Med. A-ZE  \\ 
  \hline
SpN~\cite{cao2023supernormal} GT N & 0.16	& 0.12	&0.06	& 0.10	& 0.13	& 0.11 \\
Ours GT N & 0.13	& 0.15	& 0.10 &	0.12	 & 0.06 &	0.11\\ % &  & \\

 \hline
  DiLiGenT-MV~\cite{LiZWSDT20}  & 0.22	& 0.33	& 0.08	& 0.21	& 0.25	& 0.22  \\ % &  &    \\
  PS-NeRF~\cite{psnerf} & 0.27	& 0.33	& 0.27	 &0.26	& 0.36	& 0.30\\ % &  & \\
   % PS-NeRF~\cite{psnerf} [our normals] \textcolor{red}{maybe}& x & x &  x &  x & x & x \\ % &  & \\
%  MVPSNet & x & x  &x   & x  & x & x \\
  MVAS~\cite{cao2023mvas} & 0.25	& 0.37	& 0.21	& 0.20	& 0.52	& 0.31 \\
 SpN~\cite{cao2023supernormal} & 0.19	& 0.21	& 0.21 &	0.14	& 0.22	& \textbf{0.20} \\
 RNB~\cite{BrumentRNb24} & 0.25	& 0.21	&0.31	 &0.18	& 0.27	& 0.24 \\
\hline
Ours N  & 0.29	& 0.19	& 0.17	& 0.19	& 0.29	& 0.22\\ % &  0.45& 0.31\\
Ours I & 0.25	&0.24	&0.18	&0.34	 &0.26	& 0.26\\ % &  0.45& 0.31\\
Ours I-S & 0.34	&  0.33	&  0.25	&  0.32	&  0.30 & 	0.31\\
Ours N + I  &	0.21 & 0.19	 &0.17	&0.20	&0.22	& \textbf{0.20}\\ % & 0.44 & 0.28  \\ 
 \hline
\end{tabular}
    \vspace{-0.1cm}
    \caption{Results on original \textit{dense} DiLiGenT-MV~\cite{LiZWSDT20} benchmark. For all objects we report the Chamfer distance error as well as average error on all objects. Normals are computed using the universal PS method of  \cite{hardyunips}. We evaluate four versions of our method, including using only normal (N) loss, rendering loss with (I) and without shadows (I-S) as in  Table~\ref{tab:Tab_eval_syntheticdiligent}. %named N, I, I-S and N+I namely normal loss only (N) - no rendering, rendering loss only (I) - no normals loss, rendering with no shadows I-S and combined (N+I) 
    We also include comparison with Supernormal (SpN~\cite{cao2023supernormal}) with ground truth normals (GT N).}
    \label{tab:Tab_eval_diligent}
\end{small}
\end{center}
\end{table}

\subsection{Real data}\label{sec_results_real} 
In this section we report our results on DiLiGenT-MV~\cite{LiZWSDT20} benchmark in both original \textit{dense} setup (20 views, 96 lights) and various \textit{sparse} setups. %\textcolor{blue}{Some ablation of normal vs rendering loss vs shadows is also providing showing that for most experiments the combined approach outperforms all others.}

%We first show that there are some issues with DiLiGenT-MV~\cite{LiZWSDT20} intrinsic camera parameters and/or ground truth normals. In particular, we compute the shape estimation error by using original intrinsic parameters and obtain a significantly higher error of estimated shape than on the synthetic DiLiGenT-MV benchmark. However if we take the intrinsic parameters from Reading object and use it for all objects, the error is drastically reduced (0.11mm to 0.03mm). We note that for the Reading object, the difference of 0.06mm to 0.05mm is explained by adjusting the rotation matrices with SVD.

%(we get the same parameters by fitting a 3D into ground truth normals) we recover a very similar intrinsic values (e.g \textcolor{red}{focal lengths: $fx= XX$, $fy =YY$, principal point: $u = XX$, $v =YY$, vs $fx = XX$, $fy = YY$, $u=X$, $v = Y$}). If we using these intrinsics we obtain the same error when using ground truth normals both on synthetic and real datasets. For the rest of experiments using our network we use these corrected intrinsics. \textcolor{red}{Note the other works e.g. PS-NeRF~\cite{psnerf} or MVAS~\cite{cao2023mvas} use SOME OTHER DATA - EXPLAIN THE DATA.} 

\begin{table}[t]
\setlength{\tabcolsep}{2.0pt} %3pt
\begin{center}
\begin{small}  %footnotesize
\begin{tabular}{||c| c c c c c ||c|| }%c|c||} 
 \hline
 Method & Bear & Buddha & Cow & Pot2 & Reading & Avg. \\ %& Avg. A-ZE & Med. A-ZE  \\ 
  \hline

%\hline
%SpN~\cite{cao2023supernormal} C-GT N & 0.08	& 0.06	& 0.09	& 0.05	& 0.07	& 0.07 \\ 
%Our C-GT N & 0.02	&0.06	&0.01	&0.02	&0.05	&0.03\\ 
% &  & \\
  %  PX-Net~\cite{Logothetis22} normals& x &     x &  x &  x & x & x\\ % &  & \\

%
\hline
%Ours - just silhouette - full lights & x &  x  & x  & x &  x & 	x\\ 
\hline
%Ours - [normals only] - 6 lights  & \textcolor{red}{0.58}	&  0.19	&  0.18	&  0.20	&  0.32	& 0.29\\ 
%Ours  - [normals + intensities] - 6 lights [to run]   & 0.23	& x	& 0.19	& 0.23	& x & x \\ 

SpN~\cite{cao2023supernormal} [6,6] & 0.48	 & 0.67	 & 0.27  &	0.39	 & 1.21	 & 0.61\\ 

%Ours N [6, 6]  & 0.35 &	0.51& 	0.21 &	0.83 &	0.97	& 0.57\\ 
%Ours I [6, 6]  & 0.33 &	0.30 &	0.22 &	0.35 &	0.43 &	\textbf{0.33}\\
Ours N + I [6,6]  & 0.32	& 0.35	& 0.28	& 0.33	& 0.63 & \textbf{0.38}\\ 
\hline

Brah. et al~\cite{Brahimi_2024_CVPR} [6, 75]  & 0.38	& 0.32	&0.24	& 0.29	& 0.47	& 0.34\\ 
SpN~\cite{cao2023supernormal} [6, 30] & 0.29	& 0.25	& 0.17	& 0.17	& 0.75	& 0.33\\ 
%\hline 
%Ours N [6, 30]  & 0.24 &	0.32 &	0.16 &	0.20 &	0.41 &	0.27\\ 
%Ours I [6,30]  & 0.41	& 0.28	& 0.20	& 0.34	& 0.82	& 0.41\\ 
%Ours I -S [6,30] & 0.45	&0.37	&0.21	&0.42	& 0.97	& 0.48 \\
Ours N + I [6, 30]  & 0.28	&  0.29 & 	0.16	&  0.25	&  0.25	&  \textbf{0.25}\\
\hline

SpN~\cite{cao2023supernormal} [20, 6] & 0.28	& 0.53	& 0.28	& 0.30	& 0.36	& 0.35\\ 
%Ours N [20, 6]  & 0.59	& 0.46	& 0.34	& 0.63	& 0.45	& 0.49\\ 
%Ours I  [20, 6] &0.22  &	0.29	  &0.22	  &0.32	  &0.32	 & \textbf{0.27}\\ 
Ours N + I [20, 6] & 0.25	&0.38	&0.24	&0.52	&0.33	& \textbf{0.34}\\ 
\hline

\end{tabular}
    \vspace{-0.1cm}
    \caption{Results on \textit{sparse} DiLiGenT-MV~\cite{LiZWSDT20} benchmark. We include three sparse cases (marked with [views, lights] and compare with Supernormal~\cite{cao2023supernormal} which is the best performing dense competitor. We also include the comparison with Brahimi et al~\cite{Brahimi_2024_CVPR} which was computed with 6 views and 75 lights and is comparable to our 6 views 30 lights case. %\textcolor{red}{For that case, we also able the use of normals and shadows similarly to Tables~\ref{tab:Tab_eval_syntheticdiligent} and \ref{tab:Tab_eval_diligent}. MAYBE remove actually}    
    We note that our method significantly outperform Supernormal~\cite{cao2023supernormal} (0.38mm vs 0.61mm) on the most sparse setup [6 views, 6 lights] as well as Brahimi et. al~\cite{Brahimi_2024_CVPR} (0.25mm vs 0.34mm) in [6 views, 30 lights] setup. Supernormal~\cite{cao2023supernormal} matches our performance when more views are used.} %We also 
    \label{tab:Tab_eval_diligent_sparse}
\end{small}
\end{center}
\end{table}
%We show that using the ground truth normals with the provided intrinsics and extrincs is significantly worse than using modified  parameters (i.e. using the intrinsics of Reading and performing SDV on the roation matrices; for the Reading object specifically, the 0.06 to 0.05 improvement is achived with the correction of the rotation matrices.).  

\textbf{Original (\textit{dense}) DiLiGenT-MV setup.} Our method achieves (0.2mm) state of the art results (see Table~\ref{tab:Tab_eval_diligent}), only matching to the original DiLiGenT-MV benchmark (0.22mm) and Supernormal~\cite{cao2023supernormal} (0.2mm). Note we significantly outperform most deep learning based competitors (ie. PS-NeRF~\cite{psnerf}, MVAS~\cite{cao2023mvas} and RNB~\cite{BrumentRNb24}). Also note in PS-NeRF~\cite{psnerf} the performance of various methods was compared by including the bottom (invisible) part of the object which gave misleading results of the deep learning method outperforming the classical one proposed originally with the DiLiGenT-MV~\cite{LiZWSDT20} benchmark. We believe that  the classical method performs so well as the objects have relatively simple geometry (especially the Cow where is mostly convex and has the least amount of shadows and self reflection). The reconstructed meshes are shown in Figure~\ref{fig:diligentqualitative}.

\textbf{\textit{Sparse} DiLiGenT-MV setup.} In Table~\ref{tab:Tab_eval_diligent_sparse} provide our results on various \textit{sparse} setups of DiLiGenT-MV~\cite{LiZWSDT20} benchmark. %In particular, we show that if we reduce the number of lights available (e.g. 6 lights or 30 lights) and hence decrease the quality of normals reduces and the impact of rendering increases, justifying our key contribution. In particular, in the 6 lights case, the best performance is achieved in the rendering loss only case ,as the normal prediction error is significantly higher. For 30 and 96 lights, the combined normal and intensity case achieves the best performance.
 In particular,  we include three sparse cases (marked with [views, lights]). In all cases we compare with Supernormal~\cite{cao2023supernormal} which is the best competitor on \textit{dense} DiLiGenT-MV benchmark and has code available. We also include the comparison with Brahimi et al~\cite{Brahimi_2024_CVPR} (code is not available) which was originally computed with 6 views and 75 lights which is comparable to our 6 views 30 lights case. Our method significantly outperforms Supernormal~\cite{cao2023supernormal} (0.38mm vs 0.61mm) on the most sparse setup [6 views, 6 lights] where the normal estimates are very inaccurate (see Figure~\ref{fig:intro}) and significantly outperforms Brahimi et. al~\cite{Brahimi_2024_CVPR} (0.25mm vs 0.34mm) despite having a significantly smaller number of lights in [6 views, 30 lights] setup. Supernormal~\cite{cao2023supernormal} matches (0.34 mm vs 0.35mm) our performance when more views (20) are used. Visualisation of sparse results can be found in the supplementary material.
 
It is also noteworthy that only using 6 lights and 6 views (around 2\% of the total data) only increases the total error from 0.2mm to 0.38mm signifying the need for a more challenging multi-view photometric stereo benchmark. % where the extra images would be required to achieve sub-millimeter accuracy. Additionally, a more challenging dataset would allow newer approaches to display a larger gain vs the original method, which is still marginally worse than SOTA (0.22mm vs 0.2mm). %were to be available our approach would have display a .  %\textcolor{red}{  [EXPLAIN WHY NORMALS CANT BE MADE AS GOOD AS RENDERING]}

 %Finally, we achieve slightly inferior performance to Supernormal~\cite{cao2023supernormal}, mainly because of the Pot2 object (0.2 vs 0.15mm). However given the discovered issues with the DiLiGenT-MV benchmark and code  (or the normal maps used for input) not being available   for this method we refrain from drawing on concusions for the reasons why it slightly outperforms our method. One possible explanation is there are more locally corrupted images (like the specular highlights on Bear) that are interfering with are renderer but do not pose issue to their particular normal estimation network. %Note that  SuperNormal method itself is functionally very similar to ours when no intensities (ie SuperNormal makes use of an SDF trained with normals, silhouette and Eikonal losses) are used.

%\begin{table}[t]
%\setlength{\tabcolsep}{3.0pt} 
%\begin{center}
%\begin{small}  %footnotesize
%\begin{tabular}{||c| c c c c c ||c|| }%c|c||} 
% \hline
% Method & Bear & Buddha & Cow & Pot2 & Reading & Average SE \\ %& Avg. A-ZE & Med. A-ZE  \\ 
% \hline\hline
%  \hline
 
%  Ours WACV  & x &x &	x &	x &	x & x  \\ % &  &    \\
%  Sparse PS-NeRF & x &  
%  x &  x &  x & x & x \\ % &  & \\
%\hline
%Ours - [n. only]  & x &  x & x  & x &  x & 	x\\ % &  0.45& 0.31\\
%Ours  - [n. + i.]  & x  &  x &  x  & x  & x &  x \\ % & 0.44 & 0.28  \\ 
% \hline
%\end{tabular}
    %\vspace{0.3cm}
 %   \caption{Luces stereo real }
 %   \label{tab:tabstereo}
%\end{small}
%end{center}
%\end{table}

%% file: sections/conclusions.tex
\section{Conclusions}
%\vspace{-0.2cm}

In this work we proposed a novel multi-view photometric stereo method. Unlike most MVPS methods our approach explicity leverages per-pixel intensity renderings rather than relying mainly on estimated normals. We believe such approach is required for truly applicable and robust MVPS as the estimated normals are likely to fail on complex materials or geometries. We clearly demonstrate the benefit of leveraging intensities on a synthetic and real DiLiGenT-MV benchmark and the applicability of our method on the minimal 6 lights case. 

Finally, it is important to note that improving computational efficiency has been beyond the scope of this project, however future work can improve it significantly by following strategies proposed by Supernormal~\cite{cao2023supernormal} and integrating with the NERFACC~\cite{li2023nerfacc} framework.

% and show a slight improvement on the actual benchmark highlighting a need for a more challenging multi-view photometric stereo dataset.

%% file: sections/appendix.tex
\section{Qualitative results on sparse DiLiGenT-MV~\cite{LiZWSDT20}}\label{sec:sparse_results}

\begin{figure*}[ht]
\includegraphics[width=\textwidth]{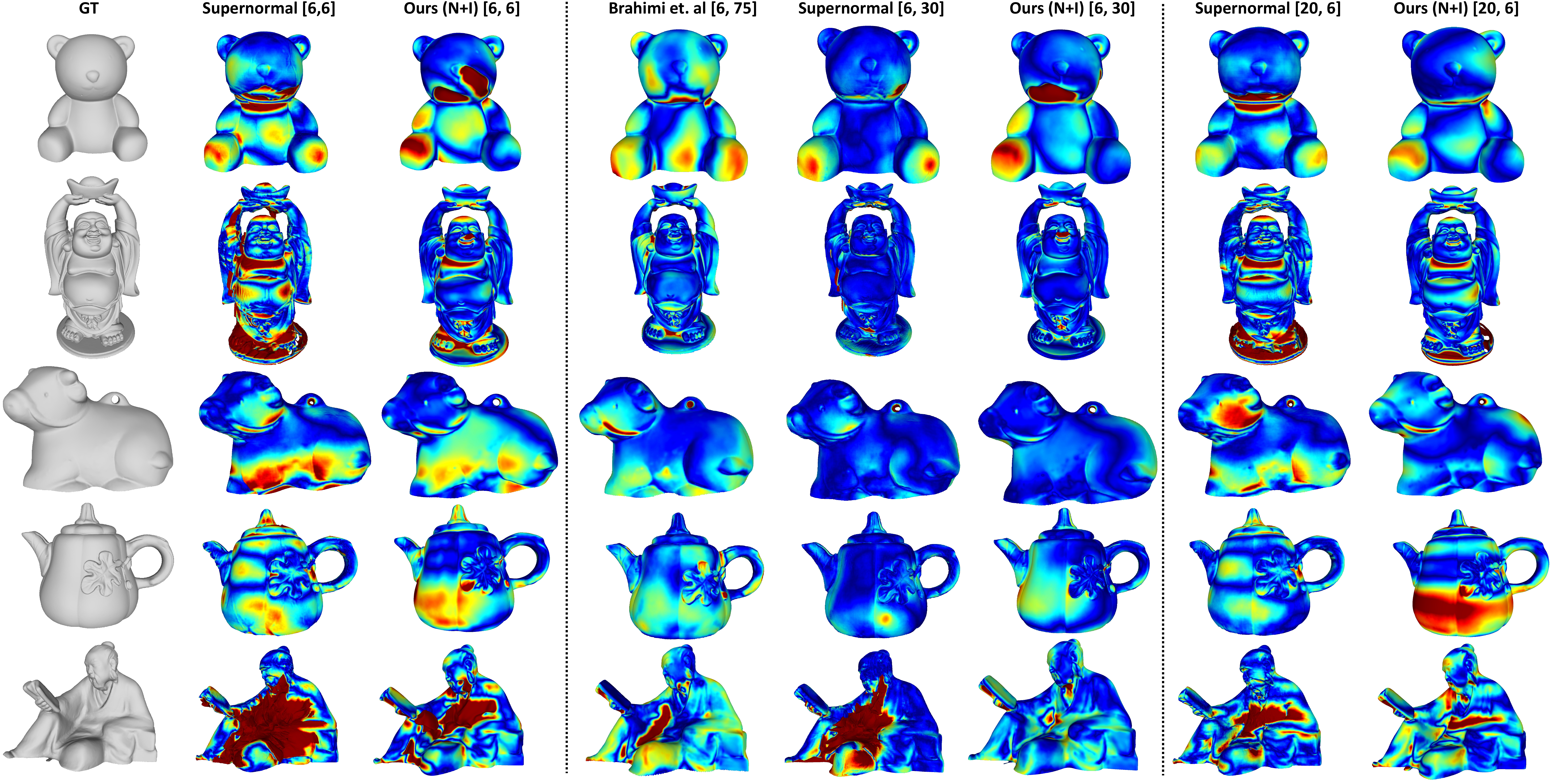}
\caption{Qualitative results on the sparse version of real DiLiGenT-MV~\cite{LiZWSDT20} benchmark. The square bracket for each case denote number of [views, lights]. For each mesh vertex, the minimum distance to the GT mesh is shown with the error bars set to red corresponding to 1mm. }
\label{fig:diligentqualitativesparse}
\end{figure*}

Qualitative results on \textit{sparse} DiLiGenT-MV~\cite{LiZWSDT20} benchmark are shown in Figure~\ref{fig:diligentqualitativesparse}. We include all 3 cases namely [6 views, 6 lights], [6 views, 30 lights] and [20 views, 6 lights]. For us, we include best version (N+I) and compare with  SpN~\cite{cao2023supernormal}, and Brahimi et al~\cite{Brahimi_2024_CVPR}, which was computed with 6 views and 75 lights which is comparable to our 6 views 30 lights case. We observe that our methods achieves consistent low error in most regions of most objects and it is thus the overall best competitor. It is notable, that  Brahimi et al~\cite{Brahimi_2024_CVPR} does not model cast shadows and thus achives high error in concavities like between the legs of  \textit{Bear}  and the inside of \textit{Reading}.

%We include three sparse cases (marked with [views, lights] and compare with SpN~\cite{cao2023supernormal} which is the best performing dense competitor. We also include the comparison with Brahimi et al~\cite{Brahimi_2024_CVPR} which was computed with 6 views and 75 lights which is comparable to our 6 views 30 lights case. We note that we significantly outperform Supernormal~\cite{cao2023supernormal} (0.38mm vs 0.61mm) on the most sparse setup [6 views, 6 lights] and also confidently outperform Brahimi et. al~\cite{Brahimi_2024_CVPR} (0.25mm vs 0.34mm) despite having a significantly smaller number of lights in [6 views, 30 lights] setup. Supernormal~\cite{cao2023supernormal} matches our performance when more views are used. \textcolor{red}{add some specific insights about qualitative results}

\section{DiLiGenT-MV data anomalies}
\label{sec:dili_prob}

This section gives additional information about identified data anomalies in DiLiGenT-MV data. First of all, we report that the rotation matrices for the Reading object do not have determinant 1 (as valid rotation matrices should). For example, for view 1 this is shown in Equation~\ref{eq:rmat}:

\begin{equation}
R_1= \begin{bmatrix}
    0.0238  &  1.0031  & -0.0137 \\
    0.4530  & -0.0230  & -0.8912 \\
   -0.8912 &   0.0150 &  -0.4533 \\
\end{bmatrix} 	\text{ with } det(R1)=1.0035 %$$
\label{eq:rmat}
\end{equation}
%\label{eq:rmat}
 %\end{equation}
In addition, we note that the intrinsic matrix is different for the Reading object than the rest with the difference being in the x axis focal length as well as the principal point as shown in  Equation~\ref{eq:kmat}:

\begin{align}
K_{Reading} &= \begin{bmatrix}
    3759.1  &  0  & 305.5 \\
    0  & 3759  & 255.5\\
   0 &   0 &  1 \\
\end{bmatrix} 	\\
K_{rest} &= \begin{bmatrix}
    3772.1  &  0  & 305.875 \\
    0  & 3759  & 255.875\\
   0 &   0 &  1 \\
\end{bmatrix}
\label{eq:kmat}
\end{align}

% Finally, We note that $\frac{3772.1}{3759.1}=1.0035$, thus the focal length difference factor is equal to the rotation scaling error so we can only speculate of exactly what the real parameters.
As we show in Table 2 of the main submission, the best performance and compatibility with the supplied GT normal maps was achieved with using the Reading intrinsics  for all objects, as well as fixing the scaling in rotation matrices with SVD. 

\begin{figure}[t]
\includegraphics[width=0.23\textwidth]{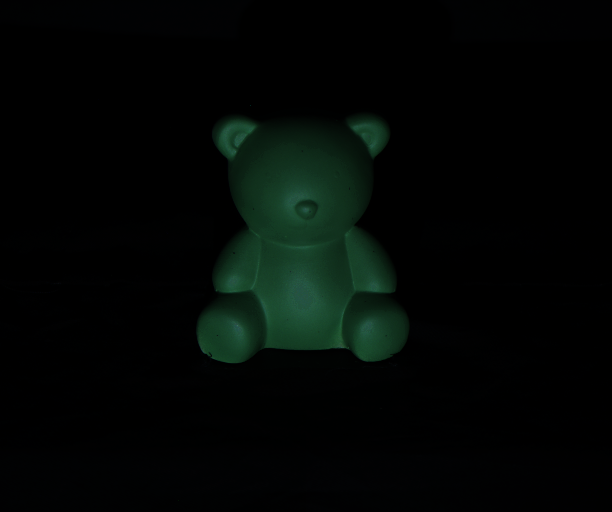}
\includegraphics[width=0.23\textwidth]{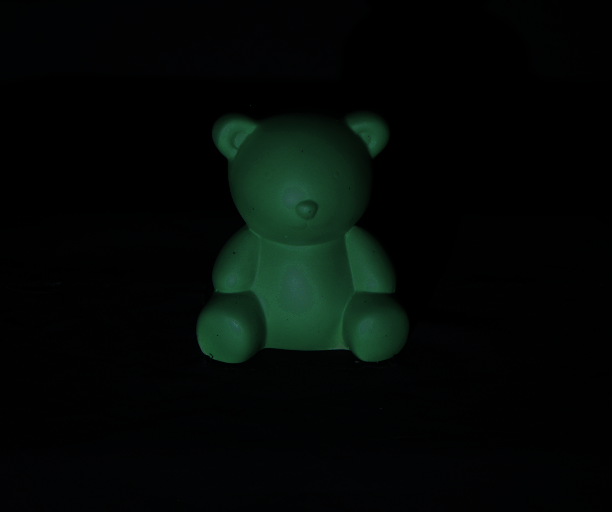}
\includegraphics[width=0.23\textwidth]{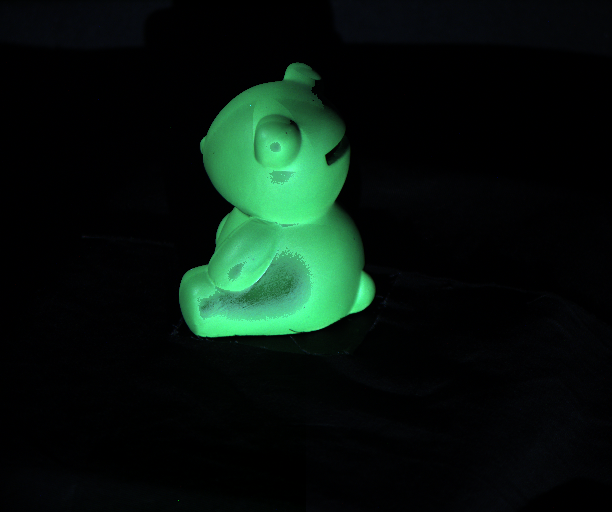}
\includegraphics[width=0.23\textwidth]{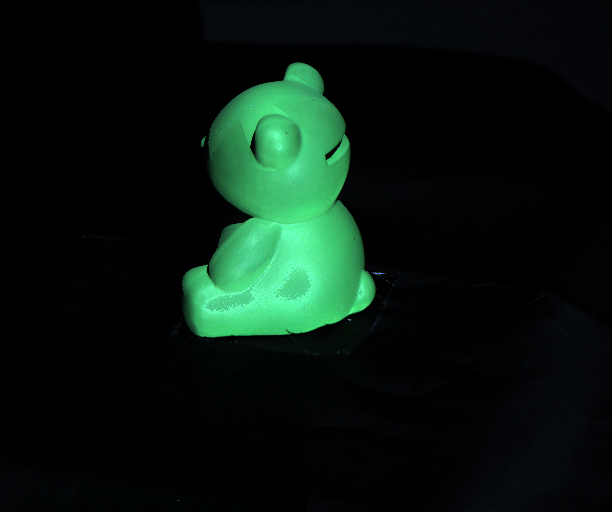}\\
\includegraphics[width=0.23\textwidth]{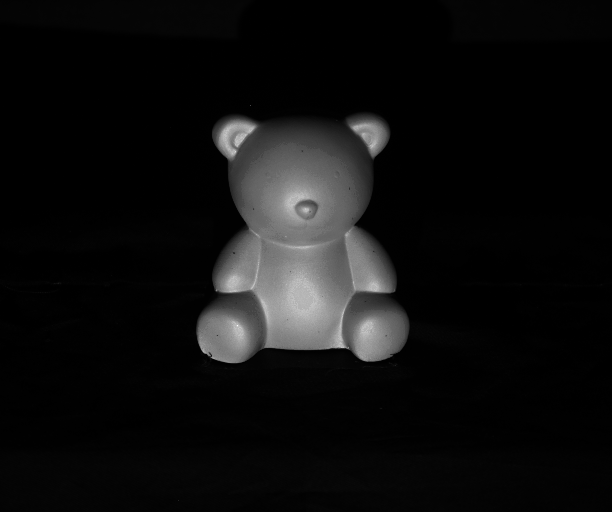}
\includegraphics[width=0.23\textwidth]{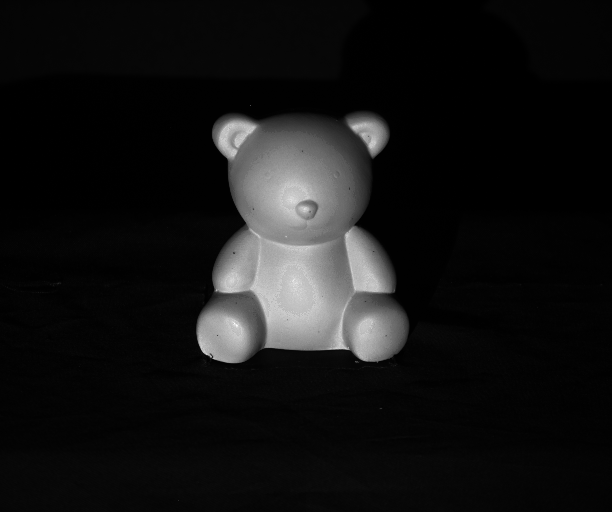}
\includegraphics[width=0.23\textwidth]{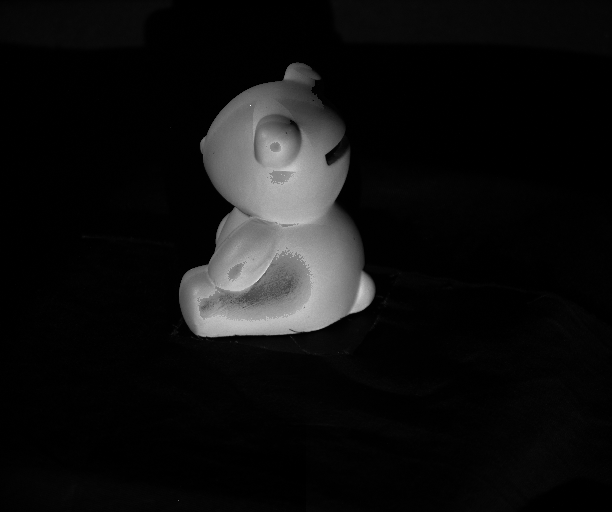}
\includegraphics[width=0.23\textwidth]{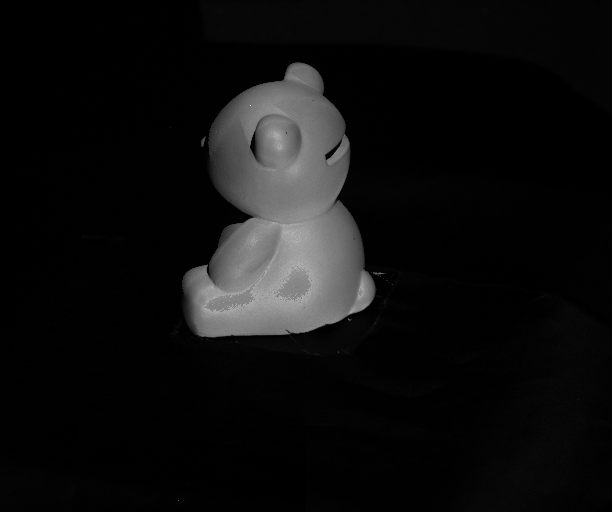}
\caption{Example of corrupted images on  DiLiGenT-MV data. From left to right view 1 lights 1 and 10, view 15 lights 48 and 64 (for the Bear object). Top row contains the original images, bottom row contains brightened up grayscale versions that make visualisation easier. It is clear that there is some data corruption around the specular highlights. }
\label{fig:corrupt}
\end{figure}

In addition, we also note that on the Bear object various images appear to be corrupted as shown in Figure~\ref{fig:corrupt}. This has been a known issue for the first view (firstly noted by Ikehata in \cite{ikehata2018cnn}) but we found corrupted images in other views. As most of the images are very dark, this is not easy to notice unless the brightness is adjusted.%; still, manually inspecting thousands of images is very time consuming so it is hard to quantify the magnitude of this issue.

\begin{figure}[t]
\includegraphics[width=0.23\textwidth]{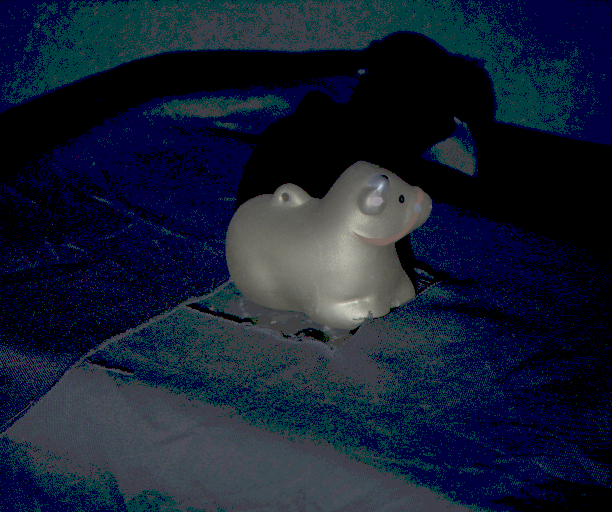}
\includegraphics[width=0.23\textwidth]{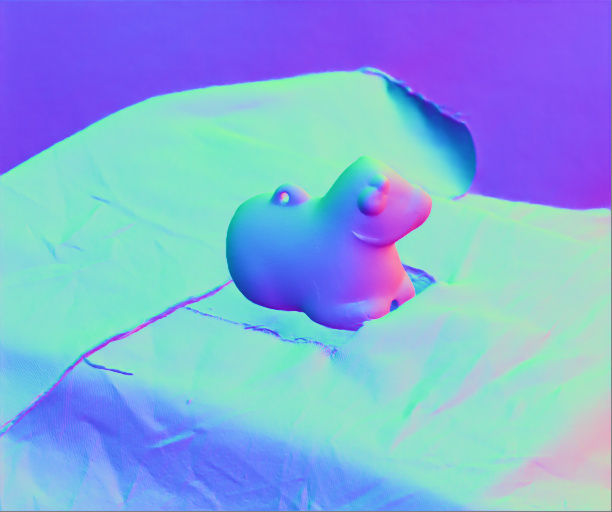}
\includegraphics[width=0.23\textwidth]{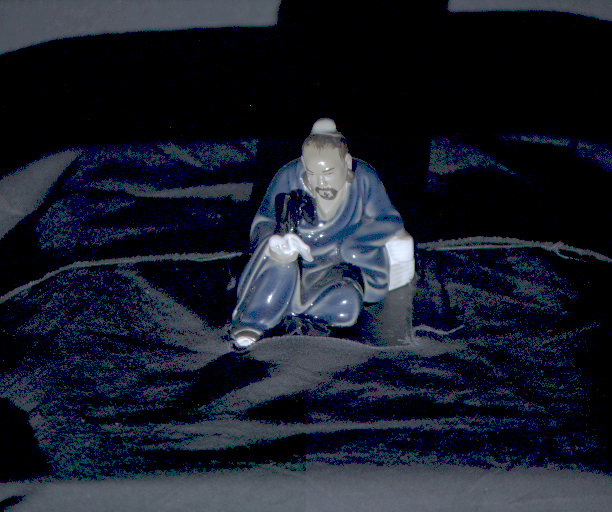}
\includegraphics[width=0.23\textwidth]{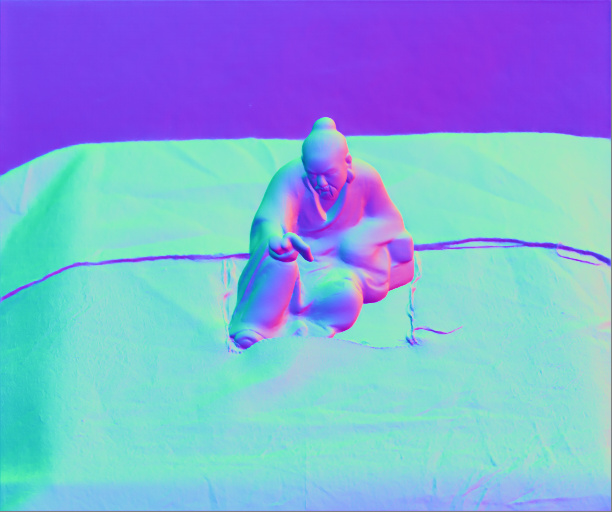}
\caption{Example of background occluding the bottom part of Cow (left) and Reading (right) objects. We show brightened average RGB image as well as full image normal maps (computed with Uni MS-PS \cite{hardyunips}) to better visualise this issue. }
\label{fig:occlusion}
\end{figure}

Finally, we note that the background seems to be occluding part of the bottom for some objects as shown in Figure~\ref{fig:occlusion}. This justifies our choise of removing the bottom 6mm of all objects for all methods in evaluating reconstruction accuracy.

